# POMDPs for Robotic Arm Search and Reach to Known Objects

Marius Silaghi  and  Jixing Zheng

Florida Institute of Technology

{msilaghi@fit.edu,jzheng2014@my.fit.edu}

We propose an approach based on probabilistic models, in particular POMDPs, to plan optimized search processes of known objects by intelligent eye in hand robotic arms. Searching and reaching for a known object (a pen, a book, or a hammer) in one's office is an operation that humans perform frequently in their daily activities. There is no reason why intelligent robotic arms would not encounter this problem frequently in the various applications in which they are expected to serve. The problem suffers from uncertainties coming both from the lack of information about the position of the object, from noisy sensors, imperfect models of the target object, imperfect models of the environment, and from approximations in computations. The use of probabilistic models helps us to mitigate at least a few of these challenges, approaching optimality for this important task.

# 1  Introduction

Robotic arms are traditionally used with automates that follow predefined trajectories, but recently they are combined with sensors to provide more intelligent functions such as abilities to open doors and grasp unknown objects. Here we address a seemingly more mundane problem of locating a known object in a partially known and bounded environment. In our problem we assume that the robotic arm has a single camera positioned in the arm. The problem is in fact challenging if we consider the need to optimize the number of movements, speed of localization, and certainty of result. The lack of stereo vision has to be compensated by taking pictures from multiple positions of the end effector, an additional challenge that adds up to the aforementioned problems.

**Applications**

One of the possible applications that we address is the robotic arm searching for objects that were misplaced on the production line due to errors (Figure 1.c). An error on the production line may place an object being processed into a wrong position and, currently, human operators are needed to stop the line and correct the situation to serve the piece in the position expected by the arms. Intelligent robotic arms could adapt to such changes automatically. Another application is in the domain of railway carriage decoupling. At railway yards, trains have to be decoupled, as carriages are selected and reorganized in new trains for different destinations. Mechanisms of various types (such as buffer-and-chain, pin-and-link, Janney, SA-3; see Figure 1.bc) are sometimes manually opened with an operation dangerous for humans, but can in the future be opened by robotic arms following methods as the one we address here [6, 7]. For example, a robotic arm can locate the pin or lever between carriages and hit/push it to decouple a train carriage which is being slowly pushed by its engine towards selection ramps at railway yards [5].

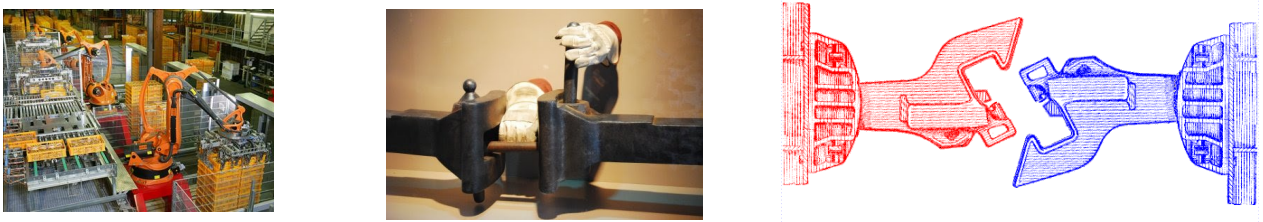

*Figure 1: Applications: (a) Sample processing line; (b) Sample pin-link coupler and (c) SA-3 holding train carriages together (Images credit to wikipedia.org).*

One of the commonly used approaches to path planning is based on search in the configuration space, where the arm is avoiding collision with elements in the environment, as well as with itself. In this work the problem of path planning including collision with other objects and among its own segments, is assumed to be solved in different module, not discussed here. In our own experiments, path planning is performed in the robot driver. Here we are concerned about the planning of a sequence of movements that maximizes the certainty of the localization of the searched object in a minimal number of image captures.

The environment, with potential positions of the object, is segmented in a hierarchical and partly overlapping structure. The search is performed in this tree, each node corresponding to a point of observation of a part of the environment. The state of the world is defined by the real position of the object. The position of the camera can be seen as either part of the state or as part of the action. The possible states are given by all possible positions of the object. With each picture, the belief as to the current state changes and the next picture point of view is planned such as to optimize this likelihood. After covering some of the POMDP theory in the next section, we continue by describing formally the problem and the proposed POMDP model. We end with discussions and conclusions.

# 2  Background

Planning problems have been addressed by robotics research for multiple decades. An important evolution of this research area consisted in the adoption of probabilistic models to represent in a scientific way the uncertainty existing in most real problems.

The source of uncertainty is constituted jointly by ignorance (e.g., concerning exact position of objects, luminosity and shape) and by the high computational complexity of known algorithms to access and process data. The ignorance is manifested not only in the lack of data but also in the incomplete modeling of physical phenomena, or in the approximations selected for modeling them.

Several approaches had been proposed to address uncertainty, including default logic, fudge factors and fuzzy logic, but the community has largely concentrated on probabilistic approaches, which are accepted as being better scientifically founded among alternatives.

Probabilistic models generally use "statements" as ontological commitments (nature of reality) and probabilities as "epistemological" commitments (possible states of knowledge), interpretable as "degrees of belief" or as "frequency", potentially describing objective properties of the world. The basic objects/statements are represented using random variables. States of the world correspond to assignments of values to these random variables.

The use of probabilistic models does not automatically reduce errors from uncertainty in reasoning except in as much as the probabilistic models do address that particular uncertainty. For example, most probabilistic models still make significant approximations concerning the actual relations (or absence of relation) between facts. Another common approximation is in discretizing time and space, and studies have addressed the convergence of these approximation towards their continuous counterparts [1].

## 2.1 Bayesian Nets

One of the most influential techniques for creating probabilistic models of phenomena is the Bayesian Network. The Bayesian Networks are graphical probabilistic models where statements (random variables) are depicted with nodes and conditional dependence relations between these concepts are displayed with directed arcs. The strength of Bayesian Networks come from the fact that not all dependence relations have to be depicted, since some of them can be inferred from others. In general, a random variable does not need to be linked to a second variable if they are independent given variables on already specified paths between them. The illustration in Figure 2 shows a simplified belief network for detecting known objects based on signals from camera interpreted as shape, color, and texture [7], in the presence of various orientations and lighting conditions.

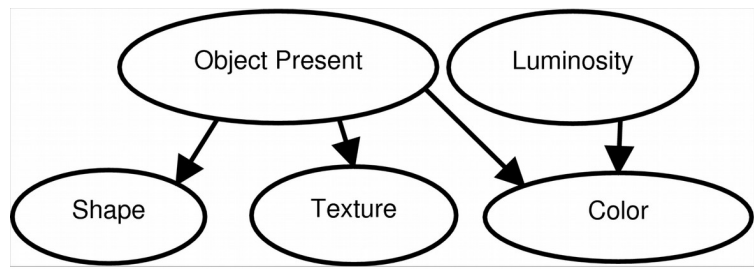

*Figure 2. Sketch Belief Network showing potential conditional dependence assumptions between variables involved in the detection of an Object, without showing conditional probability tables.*

For planning problems in environments with uncertainty in sensors or actions outcome, an alternative to continuing re-planning is to build *contingent plans* or *policies*. A policy is a mapping from each belief state of the agent into a plan to be executed in that state.

## 2.2 POMDPs

A POMDP ($\Sigma, A, T, R, \Omega, O, \gamma$) is described by a set $\Sigma$ of states, a set $A$ of actions, a set $T$ of conditional transition probabilities between states (given performed actions), a reward function $R : \Sigma \times A \rightarrow \mathbb{R}$, a set $\Omega$ of possible observations, a set $O$ of conditional observation probabilities and a discount factor $\gamma$.

Several algorithms were proposed for efficiently solving POMDPs, such as value iteration, policy iteration, point-based value iteration [2, 3, 4].

# 3  Robotic Arm Eye-in-Hand Search Problem

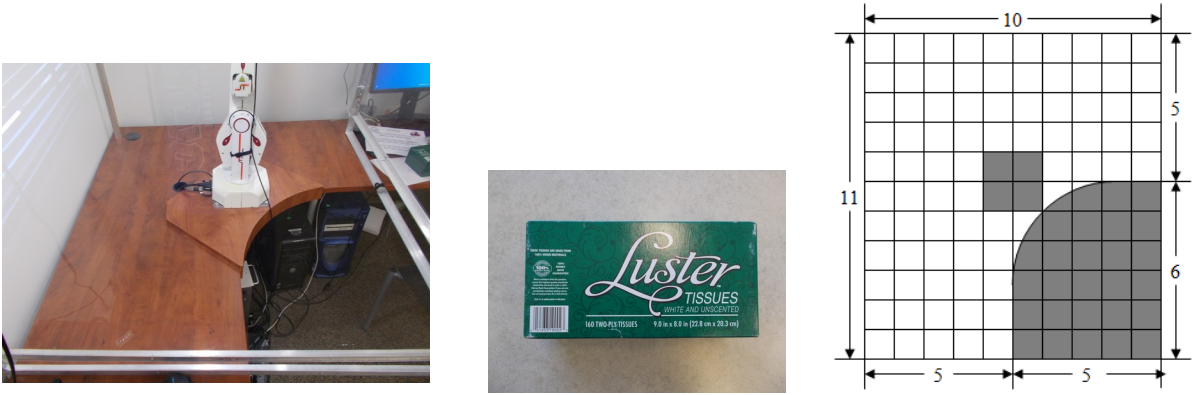

*Figure 3. Discretized Map*

A robotic arm controls an area within which it searches for a known object. For example, the area in our experiments is depicted in Figure 3.a and one of the objects (the box) is in Figure 3.b.

In our POMDP model we assume a discretized map of the working area. The problem is formally represented as a lattice $x_{i,j}$ of possible positions for the object (assumed to lay on a flat table). The object can occupy a set of 1, 2, or 3 neighboring elements of the lattice, function of its position and orientation. The object may also not be present in the environment. For the aforementioned example of the box, since it is a cuboid and the ratio of its length and width is about 2, the lattice is in Figure 3.c.

Originally the probability distribution of the position of the object is uniform across the lattice. The lattice can be analyzed from one of a set of points of observation, each of them covering a different rectangle subset in the lattice (function of the orientation, height, and location of the camera). The states seen (view) from some points of observation, subsumes the view from other points of observation (lower height). Views can also overlap. At each step, one of the possible points of observation is selected such as to maximize the expected amount of information gathered about the position of the object. We assume that capturing the object from immediate positions (1cm) corresponds to fully localizing it.

# 4 POMDP Model

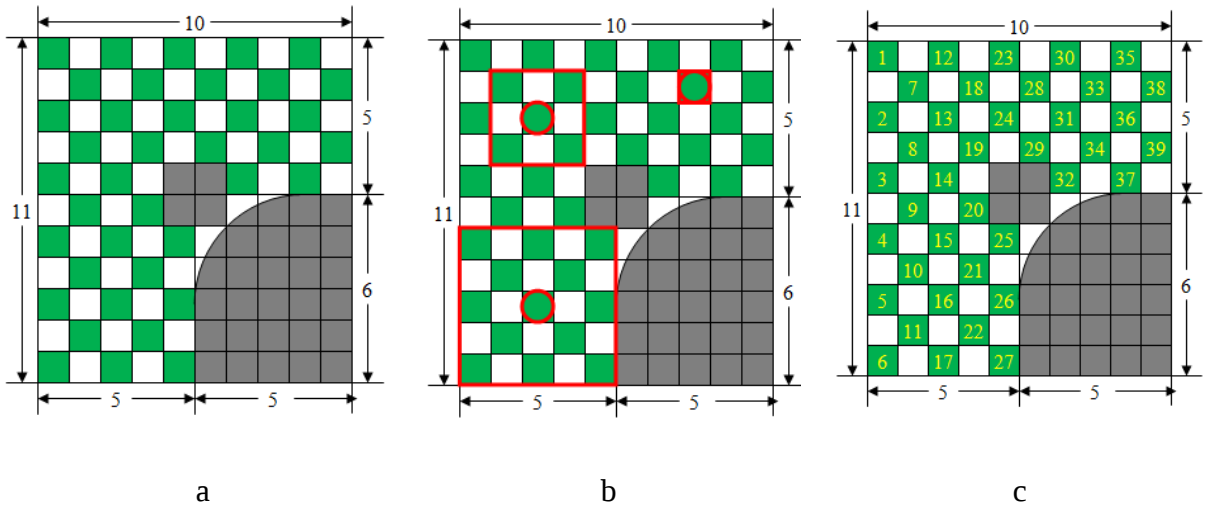

a  b  c

*Figure 4. Space reduction and partition*

The robot can view a partition of this area at a time, and the camera will move from one partition to another. The size of the block should adapt to the visual field of the camera.

**Space reduction**

Since the ratio of length and width of the green box is about 2, it can be sufficient to just observe one of the two adjacent blocks. The size of the discretized problem is compressed and the cost of computation is reduced. For example, the blocks we may move to can be reduced to the green blocks in Figure 4.a.

**Zoom of Camera**

Based on the size and the shape of the map, we define the zoom of the camera at 3 different levels (see example in Figure 4.b). Zoom 1 can only show the block that the camera is observing, so it should be an area of $1 \times 1$. Zoom 2 can show an area of $3 \times 3$ centered at the block the camera moves to, and Zoom 3 can show an area of $5 \times 5$ centered at the block the camera moves to. While our camera does not have a zoom, it is practically implemented by raising the camera at a corresponding height.

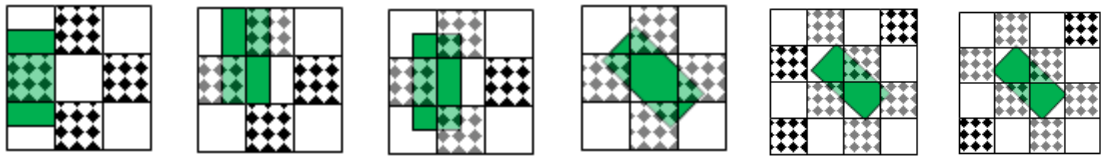

*Figure 5. End states seeing the subject*

## 4.1 States, Actions, and Transition Table

Based on the above analysis, we know that we can just move our camera from one block and zoom to another block and zoom. If we recognize part of the box at Zoom 1, then the goal is reached. We number the blocks as in Figure 4.c. At least one observed block will contain object areas. In fact, there might be more than one (in the example, at most six, see Figure 5). However, this is not addressed in our simplified model. As long as we find a block contains green area at Zoom 1, the algorithm can stop.

If the total number of blocks is *N*, we denote the statement that the object is located at block *i* with the notation $B_i, i \in \{1..N\}$.

If we do not include the position of the camera in the state, then the set *S* of all the possible states is given by Equation 1.

We denote the center block of the observed area as $C_j$, where $j = \{1, 2, \cdots, N\}$, and we denote the zoom level of the camera as $Z_k$, where $k = \{1..K\}$. With the above definition of the states, the action will consist of moving and taking a picture from a given point of view:

$$S = \{B_i | i = \{1, \ldots, N\}\} \quad (1)$$
$$A = \{snapshot[C_j, Z_k] | j = \{1, \ldots, N\}, k = \{1, \ldots, K\}\} \quad (2)$$

If we include the position of the camera in the state, then the single action is to take and analyze a snapshot while the set *S* of all the possible states is given by:

$$S = \{[B_i, C_j, Z_k] | i = \{1, \ldots, N\}, j = \{1, \ldots, N\}, k = \{1, \ldots, K\}\} \quad (3)$$
$$A = \{snapshot[C_j, Z_k] | j = \{1, \ldots, N\}, k = \{1, \ldots, K\}\} \quad (4)$$

For example, states $[B_1, C_{16}, Z_3]$, $[B_{16}, C_{31}, Z_2]$ can represent scenarios such as those in Figure 6. The initial belief distributes the probability weight equally among the states with *C* and *Z* corresponding to the current position of the camera.

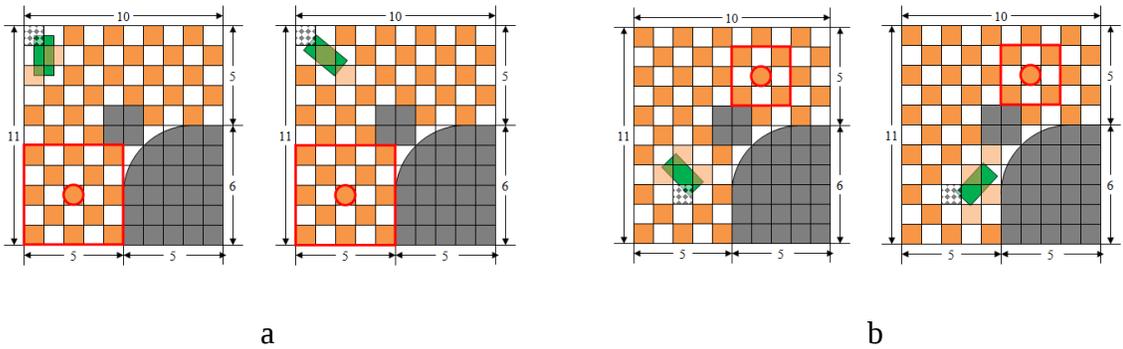

Figure 6: Sample States: a) $[B_1, C_{16}, Z_3]$, b) $[B_{16}, C_{31}, Z_2]$

**Transitions**

The transition is deterministic, since the *B* component of the state remains unchanged across transitions while the *C* and *Z* component are known whether they are part of the state or part of the action.

**Rewards**

Generally rewards are associated with states, but in our case the state desired is knowledge acquisition and the only way in which this knowledge is modeled here is via the belief function *b*. Therefore the reward of a situation can be evaluated as the mode of the belief probability function in that situation, rather than the usual *b* * *r*.

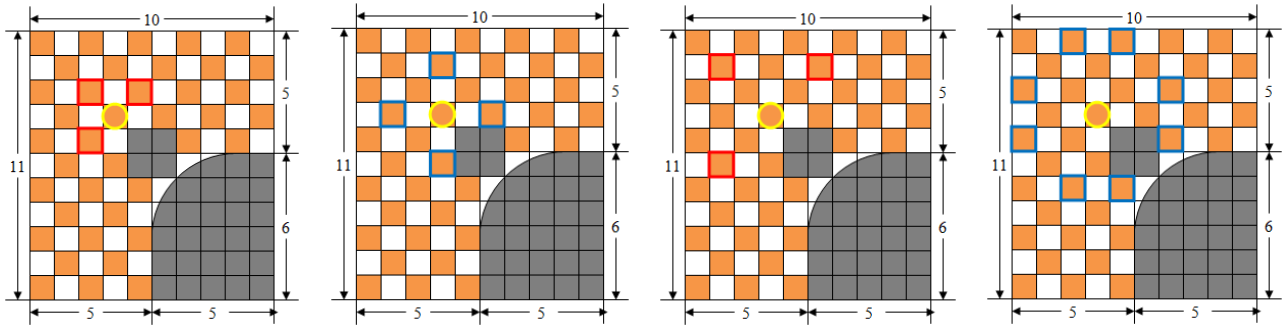

Figure 7: Distances

# Observations

### Distance between Blocks

Assuming that the side of a block has length 1, we denote with $Dist(i, j)$ the function computing the distance between the central points of block *i* and block *j*. Possible distances are $\sqrt{2}, 2, 2\sqrt{2}, \sqrt{10}$, shown in Figure 7, in a discrete sequence denoted as $\{d_i\}_{i \in \{1,2,3,4\}}$. Far away distances are represented as $d_5$ and would be used in modeling observation probabilities when snapshot views can cover larger areas. Related to the likelihood of errors, let function $Num(i, d)$ compute the number of blocks from block *i* with distance $d_i$. It is shown in Figure 8.

| $i \neq j$ | 1st nearest | 2nd nearest | 3rd nearest | 4th nearest |
|---|---|---|---|---|
| $Dist(i, j)$ | $d_1$ | $d_2$ | $d_3$ | $d_4$ |
| $Num(i, d)$ | ≤4 | ≤4 | ≤4 | ≤8 |

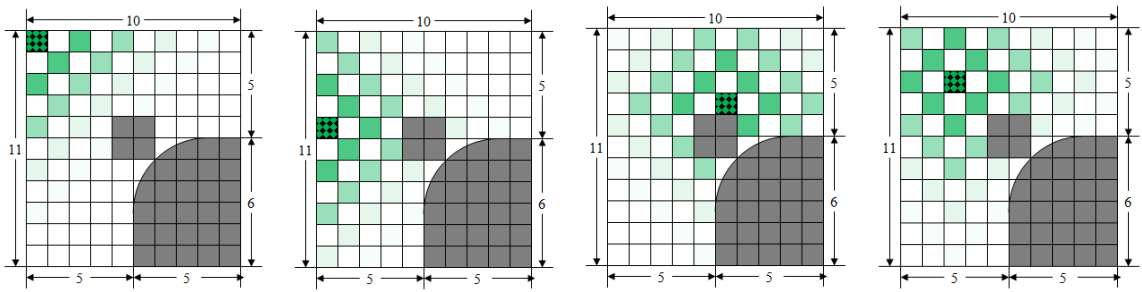

Figure 8: Expected object observation în proximity of its center

For the model discussed, there are just two possible observations. One possible observation is that the area the camera is observing contains parts of the object (denoted by $O_1$), and the other possible observation is that the area the camera is observing does not contain parts of the object (denoted by $O_2$). The certainty of the object match obtained by classifiers is expected to be distributed to blocks around the actual object, as in Figure 8.

The observation probability function $P(O|S,A)=P(O|B_i,[C_j,Z_k]) = P(O|[B_i,C_j,Z_k])$ is trained from measurements as a function of zoom and distance $f(O, Dist(i,j),k)$.

# Conclusions

In this work we show that it is possible to model a robotic arm search problem with Partially Observable Markov Decision Processes (POMDPs), where the states represent physical configurations that can encode the state of the sought knowledge (rewarding states where the camera is focusing on the unknown position of the searched object).

The obtained models allow for optimal solutions to the problem, enabling policies where large areas are inspected first before focusing on details with close-up images. While the detailed model has a low resolution, increasing the resolution can be achieved separately or in the final phases of the process.

# References

[1] Pratik Chaudhari, Sertac Karaman, David Hsu, and Emilio Frazzoli. Sampling-based algorithms for continuous-time POMDPs. In 2013 American Control Conference, pages 4604–4610. IEEE, 2013.

[2] H. Kurniawati, David Hsu, and Wee Sun Lee. Sarsop: Efficient point-based POMDP planning by approximating optimally reachable belief spaces. Robotics: Science and Systems, 2008, 2008.

[3] Joelle Pineau, Geoff Gordon, and Sebastian Thrun. Point-based value iteration: An anytime algorithm for pomdps. Technical report, Proc. of IJCAI, Mexico, 2003.

[4] Pascal Poupart, Kee-Eung Kim, and Dongho Kim. Closing the gap: Improved bounds on optimal POMDP solutions. In ICAPS, 2011.

[5] Marius C. Silaghi and Srinivasa Venkatesh. Robotic arm for searching, reaching, and unlocking train carriage couplers. In Personal Communications, 2015.

[6] Srinivasa Venkatesh and Marius C. Silaghi. Planning one eye-in-arm robot for object localization. In Proceedings of FCRAR 2015, 2015.

[7] Srinivasa Venkatesh and Marius C. Silaghi. Optimizing eye-in-arm robot for localization of known objects. In Proceedings of FCRAR 2016, 2016.